\documentclass[letterpaper, 10 pt, conference]{ieeeconf}  

\IEEEoverridecommandlockouts                              

\usepackage{graphics} 
\usepackage{epsfig} 
\usepackage{mathptmx} 
\usepackage{times} 
\usepackage{amsmath} 
\usepackage{amssymb}  
\usepackage{newtxmath} 
\usepackage{multirow} 
\usepackage{booktabs} 
\usepackage[table]{xcolor} 
\usepackage{subfigure}
\usepackage{arydshln} 
\usepackage{threeparttable} 
\usepackage{caption} 
\makeatletter
\let\NAT@parse\undefined
\makeatother
\usepackage[colorlinks, citecolor=blue, linkcolor=blue]{hyperref}
\usepackage{cleveref}

\newcommand{\insertfig}{
    \includegraphics[width=0.24\linewidth]{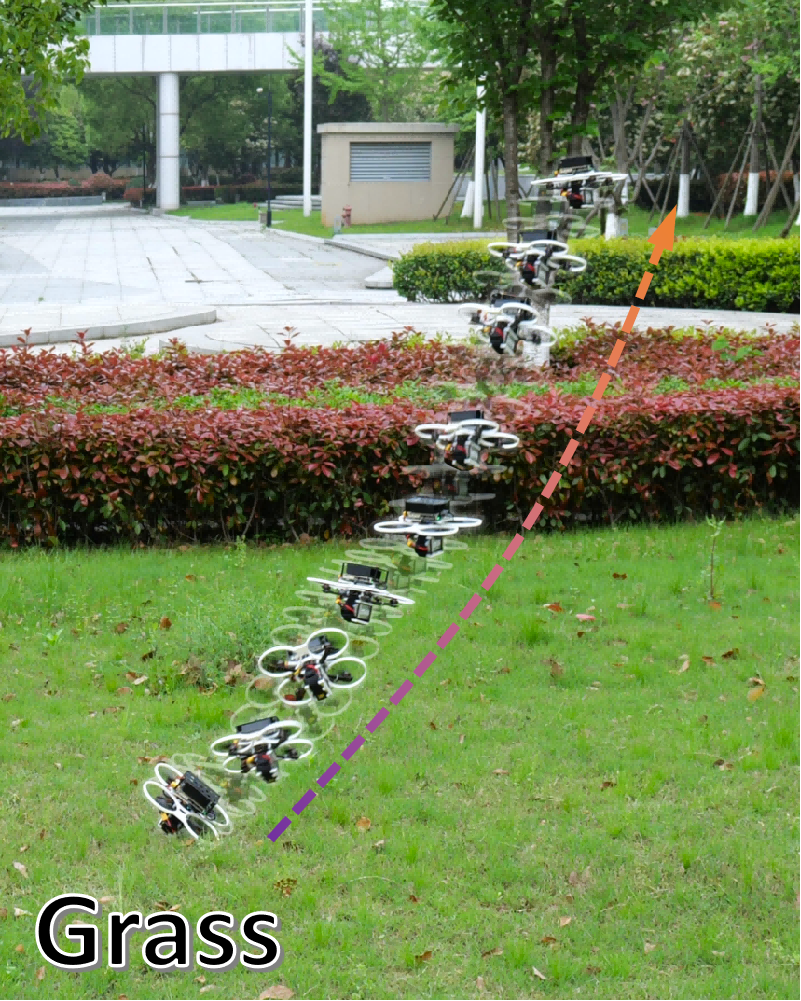}
    \includegraphics[width=0.24\linewidth]{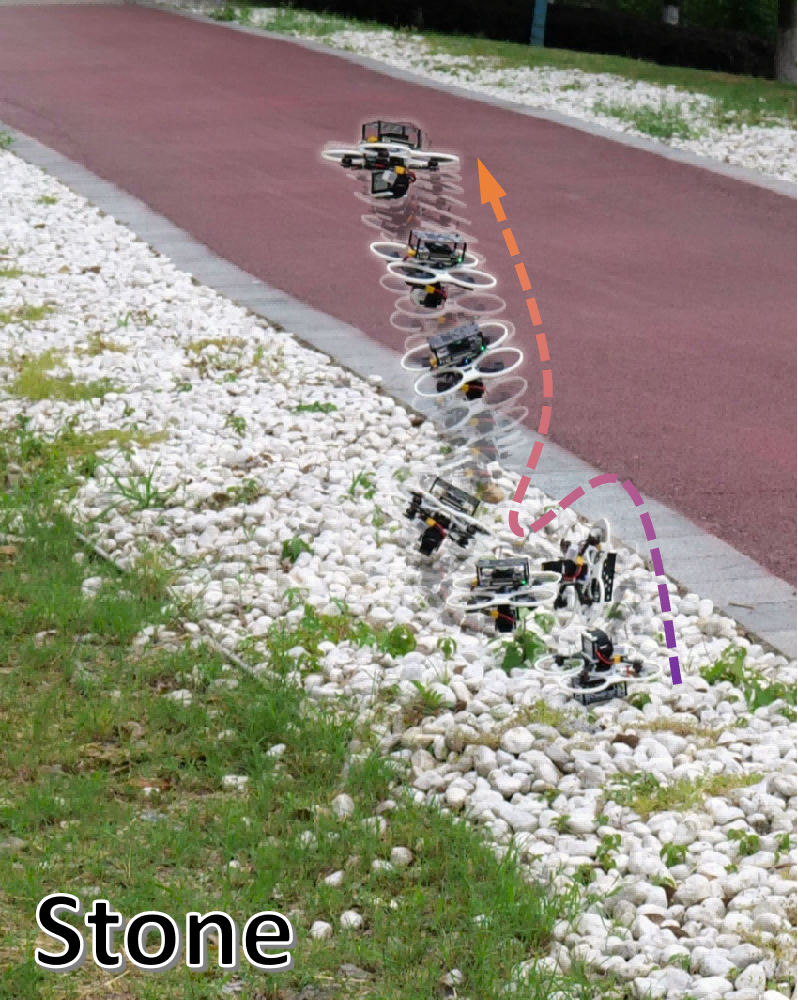}
    \includegraphics[width=0.24\linewidth]{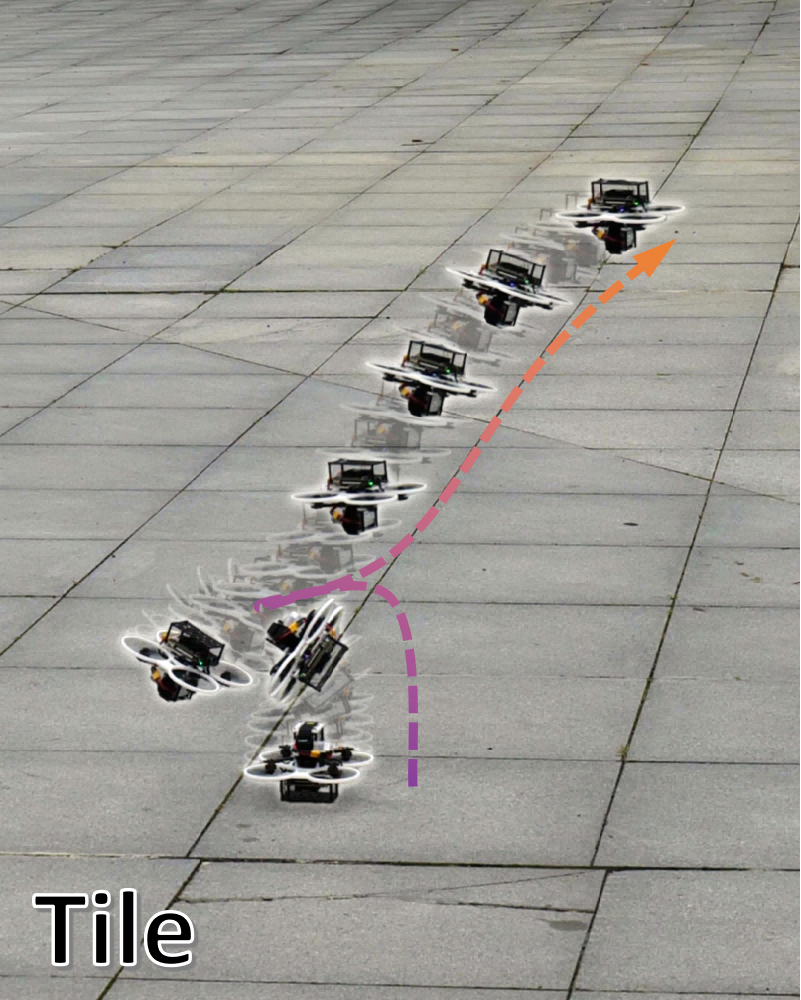}
    \includegraphics[width=0.24\linewidth]{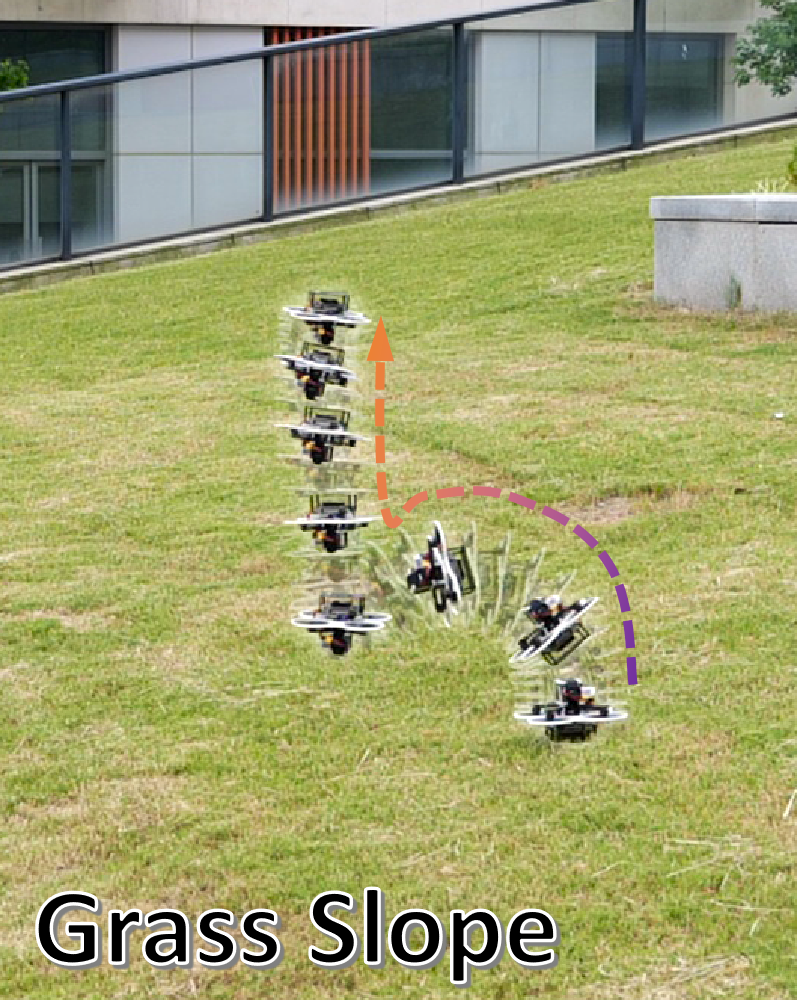}
    \captionof{figure}{Outdoor experiments. The policy can recover the quadrotor from different initial attitudes and ground surfaces to stable hover.}
    \label{fig:outdoor_exp}
    \setcounter{figure}{1}
}

\title{\LARGE \bf
Agile Fall Recovery for Quadrotors with Bidirectional Thrust via Reinforcement Learning
}

\author{Anke Zhao, Yuhang Zhong, Kenghou Hoi, Junyu Mou, Junjie Wang, \\Lijie Wang, Jialiang Hou$^\dagger$, and Fei Gao$^\dagger$
\thanks{$^\dagger$Corresponding Authors: Jialiang Hou and Fei Gao.}
\thanks{All authors are with the Institute of Cyber-Systems and Control, College of Control Science and Engineering, Zhejiang University, Hangzhou 310027, China. Jialiang Hou and Fei Gao are also with Differential Robotics, Hangzhou 311121, China.}%
\thanks{E-mail:\{AnkeZhao, jlhou25, fgaoaa\}@zju.edu.cn}%
}

\begin{document}

\makeatletter
\apptocmd{\@maketitle}{\centering\insertfig}{}{}
\makeatother

\maketitle
\thispagestyle{empty}
\pagestyle{empty}

\Crefname{figure}{Fig.}{Figs.}

\begin{abstract}

Autonomous fall recovery is a critical capability for quadrotors operating in real-world environments, where collisions or failures may leave the vehicle resting on the ground in an arbitrary attitude. This problem is challenging because recovery must be achieved under limited onboard sensing, in constrained free space, with ground contact, and in the presence of unknown disturbances. In this letter, we present an RL-based framework for autonomous fall recovery of a quadrotor from arbitrary ground attitudes to stable hover using only lightweight onboard sensors. To address severe partial observability and intermittent sensor invalidity, we train a recurrent policy within an asymmetric actor--critic architecture, leveraging an Incremental Nonlinear Dynamic Inversion (INDI) controller to track the policy output. Combined with high-fidelity simulations of motor response and optical flow, the overall training framework significantly reduces the sim-to-real gap. Simulation ablation studies validate the importance of the main design choices, while real-world experiments demonstrate zero-shot transfer and robust recovery under different initial attitudes, wind disturbances, and additional payloads. These results demonstrate that agile quadrotor fall recovery can be achieved without explicit state estimation using only limited and unreliable onboard sensing.

\end{abstract}


\section{INTRODUCTION}
Fall recovery is a key capability for quadrotors, enabling them to return to stable hover and continue their mission after collisions or failures. This capability can substantially broaden the applicability of quadrotors in real-world scenarios, such as search and rescue and infrastructure inspection, where unexpected impacts may cause the vehicle to fall.

The task of fall recovery is challenging because the vehicle must recover from arbitrary ground-resting attitudes with unknown nearby obstacles. This setting introduces constraints such as limited free space, intermittent ground contact, and possible sensor failures. The problem is further complicated by severe partial observability. During aggressive recovery maneuvers, undesirable attitudes and rapid attitude changes can lead to invalid sensor measurements, causing conventional extended Kalman filter (EKF)-based estimation pipelines to degrade or fail completely~\cite{xu2026rise}. In practical deployment, these difficulties are amplified by minimal and lightweight sensing suites, which reduce measurement redundancy and reliability. Consequently, relying on accurate, continuous state estimation is inherently fragile.

Existing work is insufficient for this setting. Prior studies on automated attitude recovery mainly focus on throw launch~\cite{faessler2015automatic, Bouman2020design}, which assumes sufficient free space and favorable initial velocity. In contrast, fall recovery requires the vehicle to generate control moments while in contact with the ground, which in turn necessitates bidirectional thrust~\cite{jothiraj2019enabling}. Although agile flight with bidirectional thrust has been demonstrated~\cite{jothiraj2019enabling, yu2020perching}, these approaches rely heavily on accurate and continuous state estimation. This assumption is difficult to satisfy using lightweight onboard sensors alone. Therefore, a robust method that can operate under severe partial observability without explicit state estimation is needed.

Inspired by recent works using reinforcement learning (RL) for agile flight without explicit state estimation~\cite{geles2024demonstrating, wu2025whole, wu2026precise}, we propose an RL-based framework for autonomous quadrotor fall recovery from arbitrary ground-resting attitudes to stable hover using only lightweight onboard sensors. Specifically, the platform uses only a downward-facing optical flow sensor and a distance sensor, which are sufficient for nominal hover while remaining lightweight and inexpensive. However, this sensor suite provides partial and unreliable measurements. To address severe partial observability, we train a recurrent policy using an asymmetric actor--critic architecture, enabling stable end-to-end learning from scratch without explicit state estimation. The learned policy outputs collective thrust and body-rate (CTBR) commands, which are tracked by an Incremental Nonlinear Dynamic Inversion (INDI) controller~\cite{smeur2016adaptive} with bidirectional thrust. The controller estimates and compensates for external torques arising from disturbances and model mismatch, thereby reducing the sim-to-real gap. Combined with high-fidelity motor-level simulation and domain randomization~\cite{tobin2017domain}, the overall pipeline enables zero-shot sim-to-real transfer to real hardware. Extensive simulation and real-world experiments validate the proposed framework, demonstrating robust recovery from different initial attitudes and across various ground surfaces under wind disturbances and payload variations.

\begin{figure*}[t]
    \vspace{5pt}
    \centering
    \includegraphics[width=1.0\linewidth]{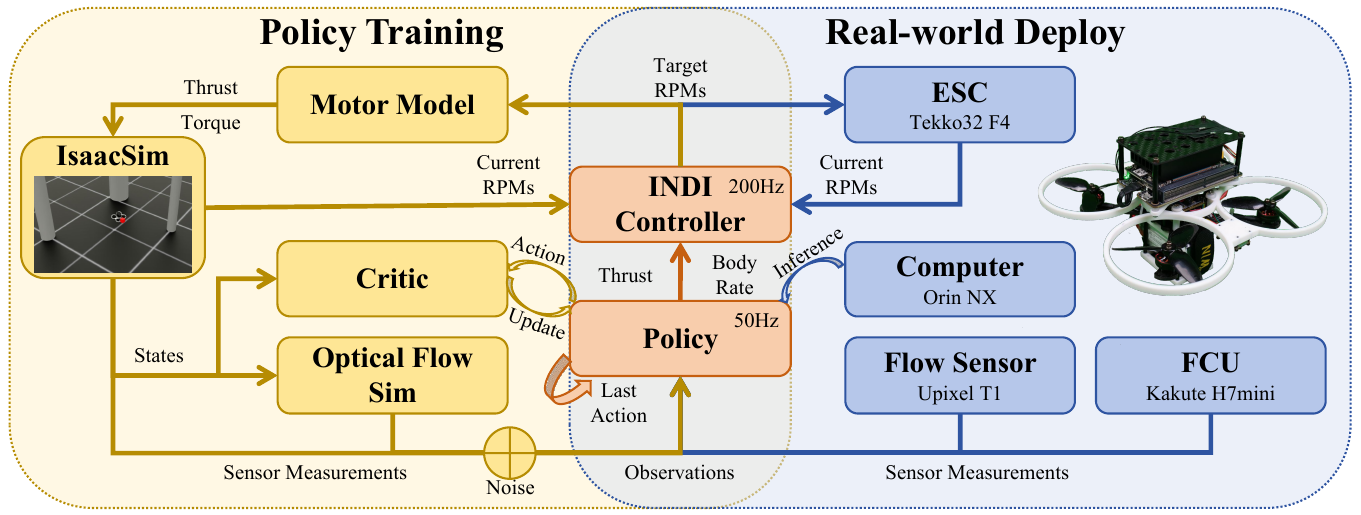}
    \caption{System overview.}
    \label{fig:overview}
    \vspace{-10pt}
\end{figure*}

The main contributions of this work are summarized as follows:
\begin{itemize}
    \item We present, to the best of our knowledge, the first method for autonomous quadrotor fall recovery from arbitrary ground-resting attitudes to stable hover using lightweight onboard sensing.
    \item We propose an RL-based fall recovery framework that combines asymmetric actor--critic architecture, INDI-based low-level controller, and high-fidelity motor-level simulation, enabling zero-shot sim-to-real transfer.
    \item We validate the proposed approach through extensive simulation and real-world experiments, demonstrating strong robustness to external disturbances and diverse ground surfaces.
    \item We open-source our code\footnote{https://anonymous.4open.science/r/Flipper} to support future research on autonomous aerial recovery.
\end{itemize}

\section{RELATED WORK}

\subsection{Fall Recovery of Legged Robots}
Fall recovery, also referred to as standing-up control, is a fundamental capability for legged robots. This problem has been extensively studied in both humanoid robots~\cite{kanehiro2003first, stuckler2006getting, haarnoja2024learning, Huang2025learning} and quadrupedal robots~\cite{lee2019robust, yang2023learning}. Early approaches typically relied on model-based controllers to track handcrafted recovery trajectories~\cite{kanehiro2003first, stuckler2006getting}. Although effective in structured settings, these methods often require substantial manual tuning and can be sensitive to disturbances and modeling errors. More recently, reinforcement learning (RL) has emerged as a powerful alternative. From methods that guide exploration using predefined motion priors~\cite{haarnoja2024learning} to policies trained entirely from scratch~\cite{Huang2025learning}, RL-based approaches have demonstrated strong robustness to model mismatch and external disturbances despite the high-dimensional action spaces of legged systems. These studies highlight the potential of RL for learning robust recovery behaviors in highly nonlinear, contact-rich settings. Motivated by this line of research, we study fall recovery for quadrotors, a problem that remains largely unexplored.

\subsection{Quadrotors with Bidirectional Thrust}
Conventional quadrotors employ fixed, unidirectional propulsion, with each propeller rotating in a prescribed direction to generate thrust for nominal flight. Although effective for standard flight, this configuration limits maneuverability and prevents recovery from certain attitudes, especially inverted configurations on the ground~\cite{jothiraj2019enabling}. Bidirectional-thrust quadrotors address this limitation by allowing propellers to reverse their rotation and generate thrust in both directions. Prior work has shown that such platforms can significantly expand the feasible flight envelope for aggressive maneuvers, including half-flips between upright and inverted flight~\cite{jothiraj2019enabling}, as well as aggressive turnaround and barrel-roll maneuvers~\cite{Wehbeh2022MPC}. Bidirectional thrust has also enabled capabilities beyond free-flight acrobatics, such as upside-down perching~\cite{yu2020perching} and perching on and detaching from walls~\cite{zou2025enabling}. In contrast to these prior studies, our work considers autonomous recovery from arbitrary ground-resting attitudes on the ground under limited onboard sensing and without explicit state estimation throughout the maneuver.

\subsection{RL-based Quadrotor Control}
RL-based methods have achieved strong performance across a broad range of quadrotor control tasks, including low-level flight stabilization~\cite{Hwangbo2017control}, aggressive acrobatic maneuvers~\cite{kaufmann2020deep}, high-speed trajectory tracking~\cite{eschmann2024learning}, and champion-level autonomous drone racing~\cite{kaufmann2023champion}. In aggressive flight regimes, RL has been shown to outperform traditional optimal control methods on certain tasks, in part because it directly optimizes task-level objectives rather than tracking predefined trajectories~\cite{song2023reaching}.

However, most existing methods either assume access to the full system state~\cite{eschmann2024learning, song2023reaching} or rely on vision-inertial odometry (VIO) for state estimation~\cite{kaufmann2023champion}. Although some prior works have demonstrated agile flight without explicit state estimation~\cite{kaufmann2020deep, geles2024demonstrating, heeg2025learning}, they still require image observations from onboard cameras. In contrast, our setting requires the quadrotor to execute an aggressive recovery maneuver from an arbitrary ground-resting ground attitude without explicit state estimation while relying only on a downward-facing optical flow sensor and a distance sensor. Furthermore, these measurements are not only noisy and partial but may also become invalid during critical phases of the maneuver, such as when the vehicle is inverted. This sensing regime creates a substantially more challenging partially observable control problem than those considered in previous RL-based quadrotor studies.

\section{METHODOLOGY}

We train a neural network policy to achieve agile fall recovery for quadrotors. The task is defined as driving the quadrotor from an arbitrary ground-resting attitude to a stable hover at a prescribed altitude directly above its initial ground location. An overview of the proposed framework is shown in~\Cref{fig:overview}. We use an asymmetric actor--critic architecture to train the policy, and directly deploy it on the real quadrotor. We use the same INDI controller in both simulation and real-world deployment, and we simulate the motor dynamics and optical flow sensor based on those of the real quadrotor.

In this section, we first describe the reinforcement learning formulation in~\Cref{sec:policy}, including the action space, observation space, reward design, and network architecture. We then present the simulation pipeline in~\Cref{sec:simulation}, including the sensor models, controller design, motor dynamics, and domain randomization strategy.

\subsection{Policy Learning}\label{sec:policy}
The policy receives measurements from a downward-facing optical-flow sensing unit, which consists of an optical-flow camera and an infrared distance sensor, as well as inertial measurements from an inertial measurement unit (IMU). Based on these observations, the neural network outputs the desired collective thrust and body rates (CTBR), a widely used control interface for aggressive quadrotor flight~\cite{kaufmann2023champion, geles2024demonstrating, wu2025whole, xu2025flying}.

We model the problem as a partially observable Markov decision process (POMDP), in which the agent interacts with the environment at discrete time steps through incomplete and noisy observations. The POMDP is defined by the tuple $(\mathcal{S}, \mathcal{O}, \mathcal{A}, p, p_0, \mathcal{R}, \gamma)$, where $\mathcal{S}$ denotes the state space, $\mathcal{O}$ the observation space, $\mathcal{A}$ the action space, $p$ the state-transition distribution, $p_0$ the initial-state distribution, $\mathcal{R}$ the reward function, and $\gamma$ the discount factor. We adopt Proximal Policy Optimization (PPO)~\cite{schulman2017proximal} because of its stability and efficiency in large-scale parallel training. Our implementation is based on RSL-RL~\cite{schwarke2025rslrl}.

\subsubsection{Action Space}
At each time step $t$, the policy outputs a four-dimensional action $\mathbf{a}_t$ consisting of the mass-normalized collective thrust $c_t$ and the desired body-rate setpoint $\boldsymbol{\Omega}^{\text{des}}_t$. An INDI controller then converts these commands into individual motor RPM setpoints, as described in~\Cref{sec:controller}. This control interface ensures the maximum agility~\cite{kaufmann2022benchmark} and does not require explicit estimation of global position or velocity during execution.

\subsubsection{Observation Space}
At time step $t$, the policy computes the action $\mathbf{a}_t$ from the observation $\mathbf{o}_t$. To improve training stability under partial observability, we adopt an asymmetric actor--critic architecture: the actor has access only to onboard sensor measurements, whereas the critic receives additional privileged ground-truth information.

Specifically, the actor receives a 19-dimensional observation vector
\[
\mathbf{o}_t = [\mathbf{o}_t^\text{ext}, \mathbf{o}_t^\text{pro}, h^{\text{des}}],
\]
where $\mathbf{o}_t^\text{ext} = [d_t, m_t^d, \mathbf{f}_t, m_t^f]$ denotes exteroceptive observations from the distance and optical-flow sensors, $\mathbf{o}_t^\text{pro} = [\dot{\mathbf{v}}_t, \boldsymbol{\Omega}_t, \mathbf{g}_t^{\mathrm{proj}}, \mathbf{a}_{t-1}]$ denotes proprioceptive observations derived from the IMU, and $h^{\text{des}}$ is the target hover height above the ground. Here, $d_t$ denotes the distance-sensor reading, and $\mathbf{f}_t = (f_t^x, f_t^y)$ is the optical-flow rate in rad/s. The binary masks $m_t^d$ and $m_t^f$ indicate whether the distance sensor and optical-flow sensor, respectively, provide valid measurements. The terms $\dot{\mathbf{v}}_t = (\dot{v}_t^x, \dot{v}_t^y, \dot{v}_t^z)$ and $\boldsymbol{\Omega}_t = (\Omega_t^x, \Omega_t^y, \Omega_t^z)$ denote the body-frame linear acceleration and angular velocity, respectively. The vector $\mathbf{g}_t^{\mathrm{proj}}$ is the normalized gravity vector projected into the body frame, and $\mathbf{a}_{t-1}$ is the previous action. All components of $\mathbf{o}_t$ can be obtained directly from onboard sensing, enabling zero-shot sim-to-real transfer without an additional state-estimation module.

In contrast, the critic receives a 33-dimensional privileged state vector
\[
\mathbf{s}_t = [\mathbf{p}_t, \mathbf{v}_t, \dot{\mathbf{v}}_t, \boldsymbol{\Omega}_t, \mathbf{g}_t^{\mathrm{proj}}, \mathbf{a}_t^{\mathrm{hist}}, \mathbf{F}_t, \mathbf{M}_t],
\]
where $\mathbf{p}_t$ denotes the relative position from the quadrotor to the target hover point, expressed in the body frame; $\mathbf{v}_t$ is the body-frame linear velocity; $\mathbf{a}_t^{\mathrm{hist}}$ contains the actions from the previous three time steps; and $\mathbf{F}_t$ and $\mathbf{M}_t$ are the random external force and moment applied to the quadrotor, as described in~\Cref{sec:domain randomization}. Because this privileged state contains information that is highly informative about future returns, the critic can estimate the value function more accurately, leading to more stable training and improved policy performance, as shown in~\Cref{sec:sim_exp}.

\subsubsection{Rewards}
We use a dense reward to guide policy learning. At each time step, the reward $r_t$ is defined as
\begin{equation}
    r_t=r^\text{hover}_t+r^\text{angle}_t-r^\text{sign}_t-r^\text{contact}_t-r^\text{smooth}_t.
\end{equation}
Each reward component is defined as follows:
\setlength{\arraycolsep}{0.0em}
\begin{eqnarray}
    \nonumber
    r^\text{hover}_t&{}={}&\lambda_1\frac{1}{1+||\mathbf{p}_t||},\\
    \nonumber
    r^\text{angle}_t&{}={}&\lambda_2\left(1-\operatorname{tanh}\left(\frac{^w\delta^z_t}{\pi}\right)\right),\\
    r^\text{sign}_t&{}={}&
        \begin{cases}
            \lambda_3,&\text{if }\operatorname{sgn}(c_t)\operatorname{sgn}(c_{t-1}) = -1,\\
            0,&\text{otherwise},
        \end{cases} \\
    r^\text{contact}_t&{}={}&
        \begin{cases}\nonumber
            \lambda_4,&\text{if contact occurs},\\
            0,&\text{otherwise},
        \end{cases} \\
    \nonumber
    r^\text{smooth}_t&{}={}&\lambda_5||\mathbf{v}_t||^2+\lambda_6||\boldsymbol{\Omega}_t||^2+\lambda_7||\mathbf{a}_t-\mathbf{a}_\text{hover}||^2\nonumber\\
            &&{+}\:\lambda_8||\mathbf{a}_t-\mathbf{a}_{t-1}||^2\nonumber.
\end{eqnarray}
Here, $^w\delta^z_t$ denotes the angle between the body-frame $z$-axis and the world-frame $z$-axis, $\operatorname{sgn}$ is the sign function, and $\mathbf{a}_{\text{hover}}=[1,0,0,0]$ is the nominal action corresponding to stable hover under ideal conditions. The hyperparameters are empirically chosen as $\lambda_1=6.0$, $\lambda_2=3.0$, $\lambda_3=2.0$, $\lambda_4=15.0$, $\lambda_5=0.05$, $\lambda_6=0.01$, $\lambda_7=0.05$, and $\lambda_8=0.05$ to balance agility and smoothness.

The terms $r^\text{hover}_t$ and $r^\text{angle}_t$ encourage the quadrotor to reach the target hover position while maintaining a near-upright attitude. Because the target is located directly above the takeoff location, $r^\text{hover}_t$ also penalizes lateral drift, even though the policy has no access to explicit position estimates. The contact penalty $r^\text{contact}_t$ discourages collisions with the ground and nearby objects, which is important for hardware safety during real-world deployment. Finally, $r^\text{sign}_t$ and $r^\text{smooth}_t$ regularize the control behavior by penalizing frequent thrust-direction reversals, large linear velocities, high angular rates, and abrupt changes in the action.
\setlength{\arraycolsep}{5pt}

\subsubsection{Network Architecture}
Because the policy does not have access to explicit state estimation, we incorporate a gated recurrent unit (GRU) with a hidden dimension of 64 to infer latent state information, such as velocity and displacement relative to the takeoff point, from the observation history. The recurrent layer also improves temporal consistency and helps smooth the resulting actions. The GRU output is then passed to a two-layer multilayer perceptron (MLP) that produces the action.

Under the asymmetric actor--critic architecture, the critic has direct access to the privileged state and therefore does not require temporal memory to estimate the value function. Accordingly, the critic is implemented as a three-layer MLP without recurrent components.

\subsection{High-Fidelity Simulation}\label{sec:simulation}
We use Isaac Lab as the training platform. To achieve high-fidelity simulation, we construct a 3D model of the real quadrotor in SolidWorks and export it to Isaac Sim. The vehicle mass, center of mass, and inertia are extracted from the CAD model and used to configure the simulator.

To reduce the sim-to-real gap, we use the same low-level controller in simulation as on the real quadrotor. Specifically, an INDI controller tracks the desired body rates and computes the desired body torques. These torques, together with the commanded collective thrust, are then mapped to individual motor thrust commands and subsequently converted to motor RPM setpoints using the propeller model. We model the motor-speed response using data collected from the real platform. Finally, the propeller model and the simulated motor RPMs are used to compute the thrust and torque generated by each rotor, which are then applied in simulation.

\begin{figure}[t]
    \vspace{5pt}
    \centering
    \includegraphics[width=1.0\linewidth]{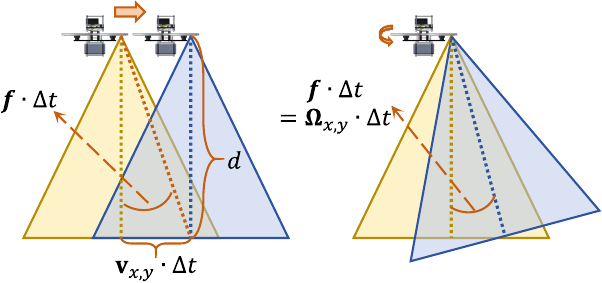}
    \caption{Illustration of optical flow induced by linear and angular motion.}
    \label{fig:flow}
    \vspace{-10pt}
\end{figure}

\subsubsection{Optical-Flow Simulation}
An optical-flow sensor estimates motion by capturing images separated by a time interval $\Delta t$ and computing the inter-frame displacement in pixels. Intuitively, this process measures how far the current image has shifted relative to the previous one. When combined with a distance sensor, the optical-flow measurement can be converted into a flow rate in rad/s. This flow rate reflects the combined effects of rotational and translational motion, as illustrated in~\Cref{fig:flow}. The theoretical flow rate is given by
\begin{equation}
    \mathbf{f}=\boldsymbol{\Omega}_{x,y}+\frac{\mathbf{v}_{x,y}}{d},
\end{equation}
where $\mathbf{v}_{x,y}$ denotes the body-frame linear velocity along the $x$- and $y$-axes, and $d$ is the measured distance to the ground.

We further compensate for the discrepancy between the IMU angular-rate measurements and the optical-flow sensor output using calibration data collected from the real quadrotor. To model sensing delay, the simulated flow measurement from $0.02$~s earlier is used as the current observation. If the flow magnitude exceeds a predefined threshold, the output is set to zero to mimic invalid readings during high-speed motion, when the inter-frame overlap becomes insufficient. We additionally inject noise whose magnitude is proportional to $||\boldsymbol{\Omega}||$ and inversely proportional to $d$. Finally, random dropout is introduced with a probability proportional to the flow magnitude to reproduce the reduced reliability of optical-flow sensing during fast motion.

\subsubsection{INDI Controller}\label{sec:controller}
To improve sim-to-real transfer, an INDI controller is used to track the desired body rates predicted by the policy. First, the desired angular acceleration is computed using a proportional controller:
\begin{equation}
    \dot{\boldsymbol{\Omega}}^\text{des}=K_{\boldsymbol{\Omega}}(\boldsymbol{\Omega}^\text{des}-\boldsymbol{\Omega}),
\end{equation}
where $K_{\boldsymbol{\Omega}}$ is the control gain. The desired control moment $\mathbf{M}^\text{des}$ is then computed as
\begin{equation}
    \mathbf{M}^\text{des} = \mathbf{M}^\text{est}+K_{\dot{\boldsymbol{\Omega}}}(\dot{\boldsymbol{\Omega}}^\text{des} - \dot{\boldsymbol{\Omega}}),
\end{equation}
where $\mathbf{M}^{\text{est}}$ is the moment estimated from the current motor RPM measurements~\cite{smeur2016adaptive}. The desired motor RPMs $\bar{\omega}_i$ are then obtained using the inverse quadrotor dynamics:
\begin{equation}
    \begin{bmatrix}
        \bar\omega_1^2\\ \bar\omega_2^2\\ \bar\omega_3^2\\ \bar\omega_4^2
    \end{bmatrix}=
    \begin{bmatrix}
        k_c & k_c & k_c & k_c \\
        -rk_c & rk_c & rk_c & -rk_c \\
        rk_c & -rk_c & rk_c & -rk_c \\
        k_m & k_m & -k_m & -k_m
    \end{bmatrix}^{-1}
    \begin{bmatrix}
        c \\ M^\text{des}_x \\ M^\text{des}_y \\ M^\text{des}_z
    \end{bmatrix},
\end{equation}
where $k_c$ and $k_m$ are the thrust and drag-moment coefficients of the propellers, and $r$ is the arm length of the quadrotor.

To support bidirectional thrust, the motor rotation direction is determined according to the sign of the desired collective thrust:
\begin{equation}
    \bar{\omega}_i=
    \begin{cases}
    \operatorname{sgn}(c)\sqrt{\operatorname{sgn}(c)\,\bar{\omega}_i^2}, &\text{if }\operatorname{sgn}c=\operatorname{sgn}\bar{\omega}_i^2,\\
    \operatorname{sgn}(c)\,\omega_\text{idle}&\text{otherwise},\\
    \end{cases}
\end{equation}
where $\omega_\text{idle}$ is the minimum idle RPM. Compared with assigning the motor direction directly from $\operatorname{sgn}\bar{\omega}_i^2$, this strategy reduces the risk of motor overheating caused by frequent reversals in rotation direction.

\subsubsection{Motor Responses}
Following Aerial Gym~\cite{kulkarni2025aerial}, the motor dynamics are modeled as a first-order system with different time constants for acceleration and deceleration. Let $\omega_i$ denote the current motor speed and $\bar{\omega}_i$ the commanded motor speed. To simulate the communication delay between the onboard computer and the flight controller, a one-step delay is applied such that the target speed at time $t$ is given by $\bar{\omega}_{i,t-1}$. For simplicity, in the equations below, $\bar{\omega}_i$ denotes the delayed commanded speed.

The effective time constant $\tau_i$ of each motor is determined by
\begin{equation}
    \tau_i =
    \begin{cases}
        \tau_{\text{inc}}, &|\bar{\omega}_i|-|\omega_i|>0\text{ and }
        \operatorname{sgn}(\bar{\omega}_i)=\operatorname{sgn}(\omega_i),\\
        \tau_{\text{dec}}, &\text{otherwise}.
    \end{cases}
\end{equation}
The continuous-time motor model is
\begin{equation}
    \dot{\omega}_i = \operatorname{clip}\left(\frac{\bar{\omega}_i-\omega_i}{\tau_i}, -\dot{\omega}_\text{max}, \dot{\omega}_\text{max}\right),
\end{equation}
where $\dot\omega_\text{max}$ is the maximum motor slew rate. In implementation, we integrate the motor dynamics using a fourth-order Runge--Kutta (RK4) scheme to balance numerical accuracy and computational cost. This model captures command latency, asymmetric spin-up and spin-down responses, direction changes, and actuator slew-rate limits while maintaining stable rotor-speed trajectories during simulation.

\begin{table}[t]
    \vspace{7pt}
    \centering
    \renewcommand{\arraystretch}{1.3}
    \setlength{\tabcolsep}{13pt}
    \caption{Domain randomization parameters.}
    \label{tab:domain randomization}
    \rowcolors{2}{white}{gray!10}
    \begin{tabular}{ll}
        \toprule
        \textbf{Term} & \textbf{Value} \\
        \midrule
        Thrust coefficient $k_c$    & $\mathcal{U}(0.9,1.1)\times k_c$ \\
        Moment coefficient $k_m$    & $\mathcal{U}(0.9,1.1)\times k_m$ \\
        Inertia                     & $\mathcal{U}(0.9,1.1)\times$ nominal inertia \\
        Random force $\mathbf{F}$   & $\mathcal{U}(-0.1,0.1)\times$ vehicle weight \\
        Random moment $\mathbf{M}$  & $\mathcal{U}(-0.002,0.002)$ N$\cdot$m \\
        Angular velocity $\boldsymbol{\Omega}$          & $\boldsymbol{\Omega}+\mathcal{N}(0,0.2)$ rad/s \\
        Angular acceleration $\dot{\boldsymbol{\Omega}}$& $\dot{\boldsymbol{\Omega}}+\mathcal{N}(0,0.5)$ rad/s$^2$ \\
        Linear acceleration $\dot{\mathbf{v}}$          & $\dot{\mathbf{v}}+\mathcal{N}(0,0.5)$ m/s$^2$ \\
        Projected gravity $\mathbf{g}^{\text{proj}}$    & $\mathbf{g}^{\text{proj}}+\mathcal{N}(0,0.1)$ m/s$^2$\\
        \bottomrule
    \end{tabular}
    \vspace{-10pt}
\end{table}

\subsubsection{Domain Randomization}\label{sec:domain randomization}
To bridge the gap between simulation and reality, we apply domain randomization~\cite{tobin2017domain} to improve policy robustness. The randomized parameters are listed in~\Cref{tab:domain randomization}. We apply randomization to three aspects: quadrotor dynamics, observations, and external disturbances.

Importantly, observation noise is added only to the actor inputs $\mathbf{o}_t$ and to the feedback signals used by the controller. In contrast, the critic receives noise-free privileged states $\mathbf{s}_t$. These randomizations emulate model mismatch, sensor imperfections, and unmodeled disturbances, encouraging the policy to infer appropriate actions from noisy observations rather than memorize a fixed open-loop action sequence.

\section{SIMULATION EXPERIMENTS}\label{sec:sim_exp}

To evaluate the effectiveness of the proposed design and the robustness of the learned policy, we first conduct a set of ablation studies in simulation. The experimental setup, including the evaluation metrics and ablation variants, is described in~\Cref{sec:sim_exp_setup}, and the corresponding results are presented in~\Cref{sec:sim_exp_results}.

\subsection{Experimental Setup}\label{sec:sim_exp_setup}
\subsubsection{Evaluation Metrics}
Following the fall-recovery task definition, we consider the following evaluation metrics:
\begin{itemize}
    \item \textbf{Success rate} $E_\text{succ}$: Let $\mathbf{p}_h$ and $\mathbf{p}_h^\text{des}$ denote the projections of the quadrotor position and the target position onto the horizontal plane, respectively. Let $h_w$ and $h^\text{des}$ denote the quadrotor height and the target height. With thresholds $\epsilon_v=0.1$~m and $\epsilon_h=0.5$~m for the vertical and horizontal errors, respectively, an episode is considered successful if $|h_w-h^\text{des}|<\epsilon_v$ and $\|\mathbf{p}_h-\mathbf{p}_h^\text{des}\|<\epsilon_h$ hold at the end of the episode.
    \item \textbf{Drift distance} $E_\text{drift}$: The maximum horizontal distance between the quadrotor and the target position during the episode. This metric reflects the risk of collision with unknown nearby obstacles.
    \item \textbf{Recovery time} $E_\text{time}$: The time required to reach stable hover, reflecting task efficiency. The quadrotor is considered to have reached stable hover if $|h_w-h^\text{des}|<\epsilon_v$ and $\|\mathbf{p}_h-\mathbf{p}^\text{des}_h\|<\epsilon_h$ hold continuously for $0.5$~s.
    \item \textbf{Smoothness}  $E_\text{smth}$: The sum of the squared differences between consecutive actions over an episode. This metric quantifies the smoothness of the policy output.
\end{itemize}

\subsubsection{Baselines}
To evaluate the effectiveness of key design choices, including the asymmetric actor--critic architecture, the network structure, the low-level controller, and the action modality, we compare the proposed method against the following ablated variants:
\begin{itemize}
    \item \textbf{Symmetric actor--critic}: A baseline with a symmetric actor--critic architecture, in which the critic receives the same observations $\mathbf{o}_t$ as the actor and uses the same network architecture.
    \item \textbf{Recurrent critic}: A baseline in which the critic uses the same recurrent network architecture as the actor while still receiving privileged states $\mathbf{s}_t$.
    \item \textbf{MLP actor}: A baseline in which the actor is implemented as an MLP without a GRU layer.
    \item \textbf{Proportional controller}: An ablation of the INDI controller in which the desired angular accelerations are tracked using a proportional controller.
    \item \textbf{RPM action}: A baseline in which the action space consists of the desired RPMs of the four motors.
\end{itemize}

All variants are trained with 2048 parallel environments for 3000 epochs, corresponding to approximately 6 million simulation steps. Training takes about 30 minutes on a desktop computer equipped with an Intel Core i7-13700KF CPU and an NVIDIA GeForce RTX 4060 Ti 16G GPU. Each trained policy is evaluated over 5000 episodes.

\begin{figure}
    \vspace{3pt}
    \centering
    \includegraphics[width=1.0\linewidth]{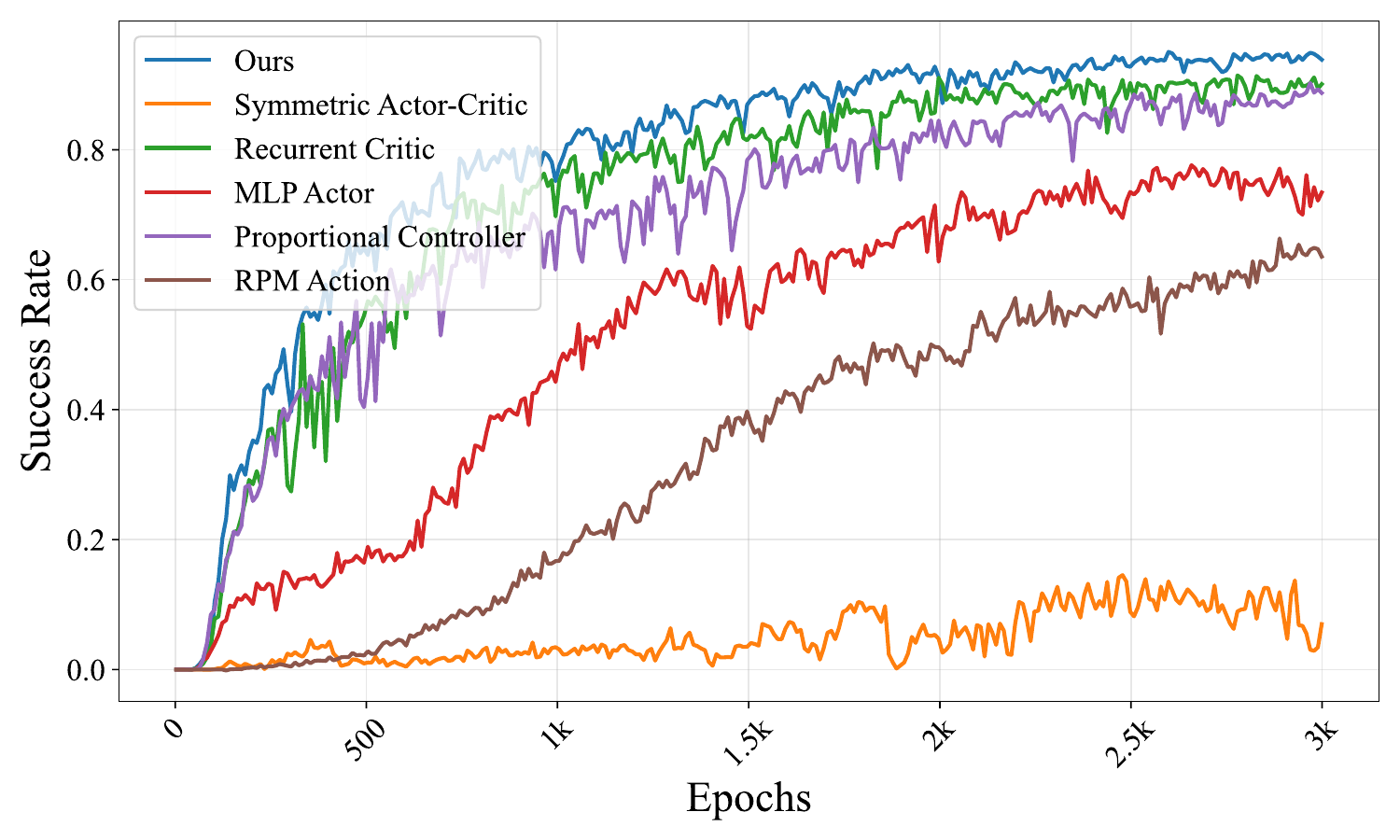}
    \caption{Success rate evolution of all variations during training.}
    \label{fig:sim_ablation}
\end{figure}

\begin{table}
    \centering
    \renewcommand{\arraystretch}{1.3}
    \setlength{\tabcolsep}{3.5pt}
    \caption{Main Simulation Results}
    \label{tab:sim_result}
    \rowcolors{2}{white}{gray!10}
    \begin{threeparttable}
    \begin{tabular}{lcccc}
        \toprule
        \textbf{Method} & $E_\text{succ}\uparrow$ & $E_\text{drift}\downarrow$ & $E_\text{time}\downarrow$ & $E_\text{smth}\downarrow$ \\
        \midrule
        \textbf{Ours}           & $\mathbf{95.04}$  & $\mathbf{0.35}\pm0.23$    & $\mathbf{1.26}\pm0.53$    & $\mathbf{0.26}\pm0.16$ \\
        Recurrent Critic        & $89.32$           & $0.44\pm0.33$             & $1.33\pm0.55$             & $0.33\pm0.30$ \\
        Proportional Controller & $89.14$           & $0.43\pm0.30$             & $1.32\pm0.51$             & $0.41\pm0.32$ \\
        MLP Actor               & $86.56$           & $\mathbf{0.35}\pm0.19$    & $1.48\pm0.63$             & $0.35\pm0.13$ \\
        RPM Action              & $69.18$           & $0.48\pm0.31$             & $1.85\pm0.84$             & $N/A$\tnote{*}\\
        Symmetric Actor--Critic & $7.00$            & $1.16\pm0.43$             & $2.06\pm1.01$             & $0.89\pm0.85$ \\
        \bottomrule
    \end{tabular}
    \begin{tablenotes}
        \footnotesize
        \item[*] With a different action space, the smoothness of RPM action cannot be directly compared with other settings.
    \end{tablenotes}
    \end{threeparttable}
    \vspace{-10pt}
\end{table}

\subsection{Experimental Results}\label{sec:sim_exp_results}
The proposed policy successfully recovers the quadrotor to a stable hover while maintaining limited drift, as shown in~\Cref{fig:sim_ablation}. The statistical results in~\Cref{tab:sim_result} further demonstrate the advantages of the proposed design in terms of success rate, safety, efficiency, and smoothness.

From the perspective of network architecture, the symmetric actor--critic variant fails to learn effectively under partial observability, highlighting the importance of the asymmetric actor--critic design. By providing the critic with privileged information, the asymmetric formulation yields more informative value estimates, thereby improving both training stability and final policy performance. The recurrent-critic variant performs slightly worse than the proposed method across all metrics, suggesting that recurrent memory is unnecessary for the critic when privileged full-state information is already available. The MLP-actor variant achieves a comparable drift distance to the proposed method, but at the cost of slower recovery and less smooth control. This highlights the importance of the GRU layer in filtering noisy sensory inputs and producing smooth actions based on temporal information.

From the perspective of control design, the proportional-controller variant underperforms the proposed method on all metrics. This suggests that the INDI controller better compensates for disturbances induced by domain randomization and random external torques, thereby simplifying the learning problem for the high-level policy. Instead of having to account for such effects purely through observation history, the policy can rely on a more robust low-level tracking module. Finally, the RPM-action variant performs substantially worse than the CTBR-action formulation. This degradation is likely due to the highly nonlinear quadrotor dynamics, especially under bidirectional thrust, as well as the difficulty of compensating for model mismatch and external disturbances under partial observability when the policy directly commands motor speeds.

\begin{figure*}[t]
    \centering
    \subfigure[Lean to the right]{\includegraphics[width=0.24\linewidth]{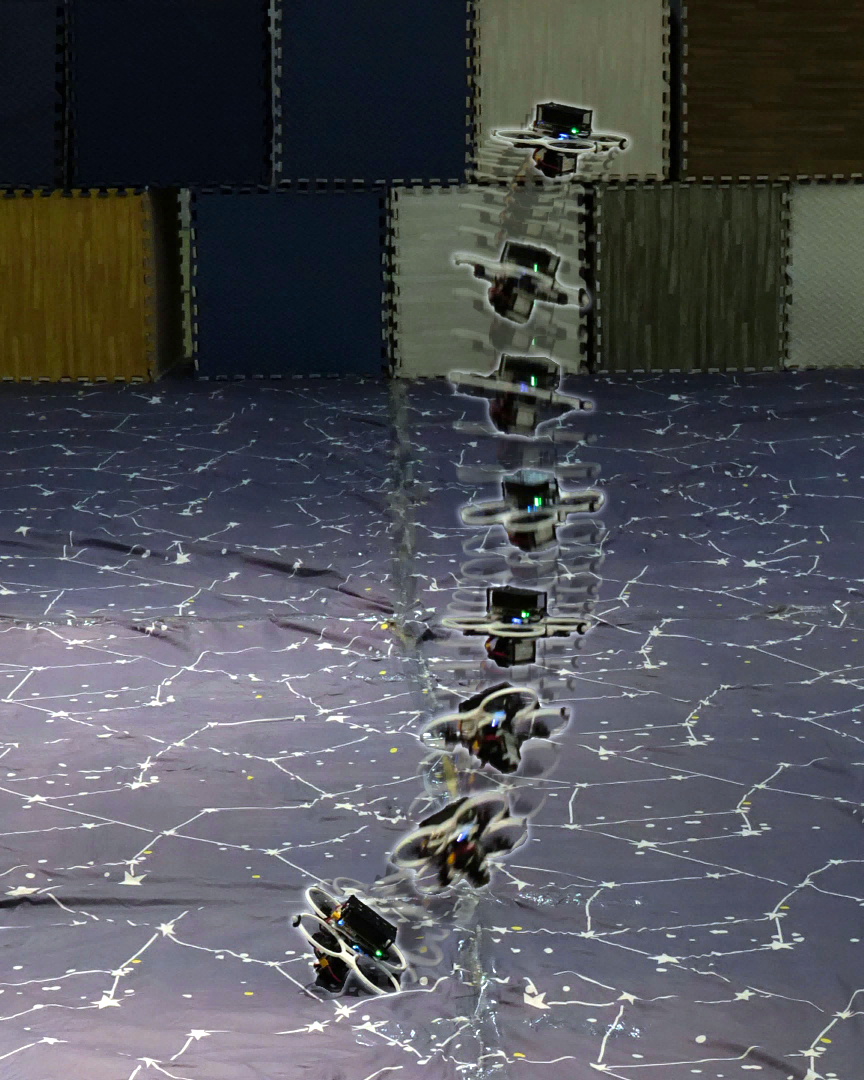}}
    \hfil
    \subfigure[Inverted]{\includegraphics[width=0.24\linewidth]{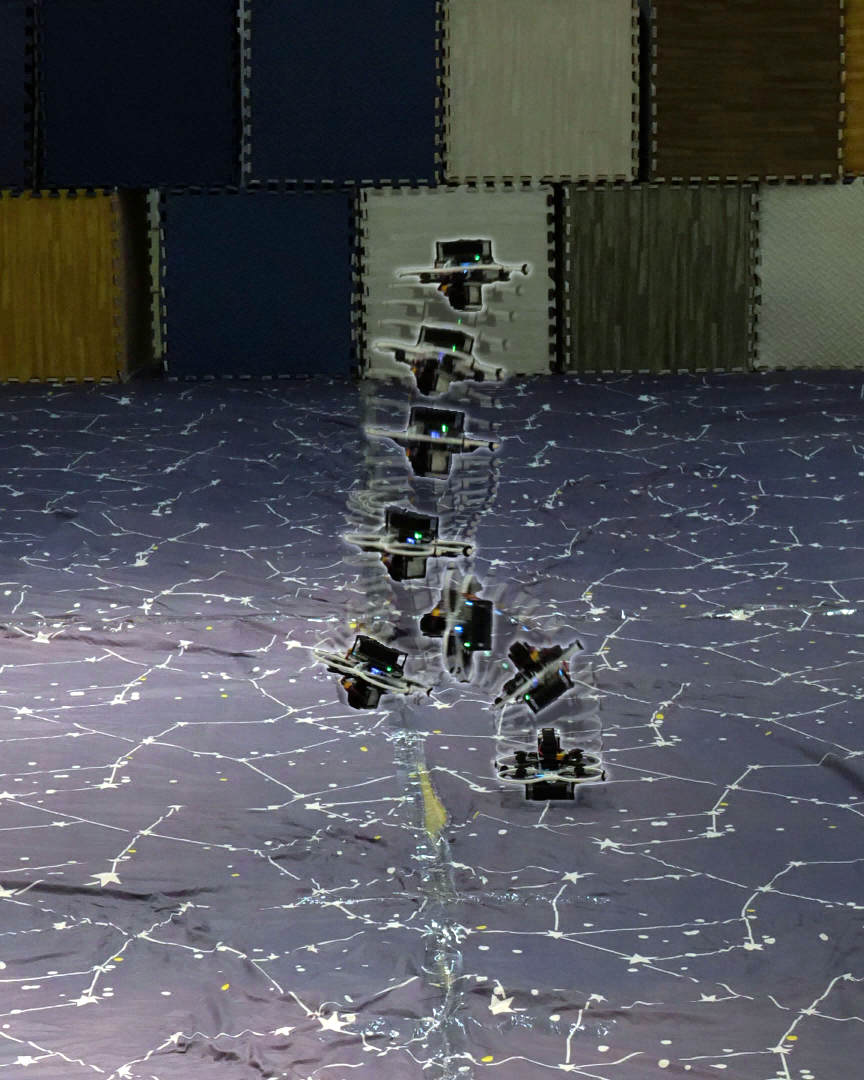}}
    \hfil
    \subfigure[With wind disturbance]{\includegraphics[width=0.24\linewidth]{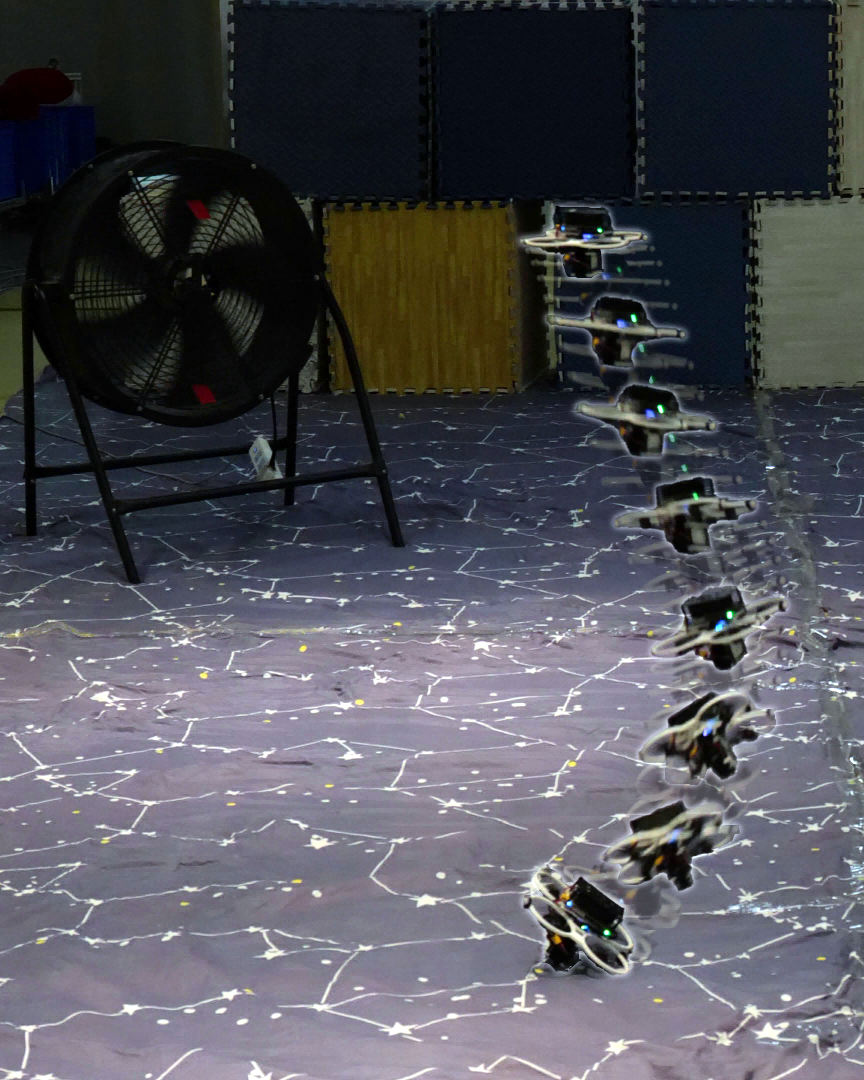}\label{fig:wind}}
    \hfil
    \subfigure[With additional weight]{\includegraphics[width=0.24\linewidth]{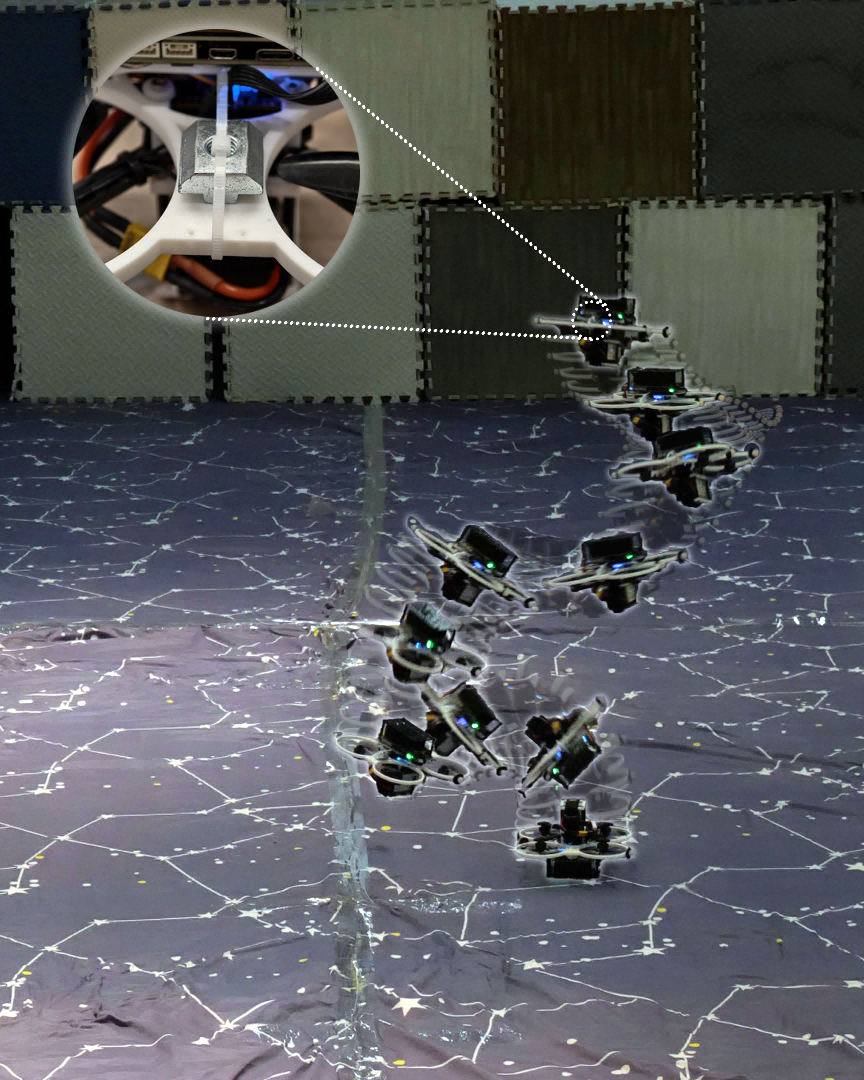}\label{fig:weight}}
    \caption{Indoor experiments.}
    \label{fig:indoor_exp}
    \vspace{-10pt}
\end{figure*}

\section{REAL-WORLD EXPERIMENTS}

To further evaluate the robustness of the learned policy and its sim-to-real transfer performance, we conduct experiments in both indoor and outdoor environments. We first describe the experimental platform and implementation details in~\Cref{sec:real_exp_system}, and then present the evaluation results in~\Cref{sec:real_exp_results}.

\subsection{Implementation Details}\label{sec:real_exp_system}
The quadrotor platform used in the real-world experiments is shown in~\Cref{fig:overview}. The vehicle is equipped with a Upixel T1 optical-flow module, which provides optical-flow-rate and distance measurements. We use the same low-level controller described in~\Cref{sec:controller} to convert the policy output into desired motor RPMs and then apply a quadratic throttle curve to map the desired RPMs to normalized throttle commands. The platform uses a Holybro Kakute H7 Mini flight control unit\footnote{https://holybro.com/products/kakute-h7-mini} running modified ArduPilot firmware\footnote{https://ardupilot.org/}, which directly transmits throttle commands to the electronic speed controllers (ESCs)~\cite{yang2025ground}. The ESCs are operated in 3D mode to enable bidirectional thrust. Policy inference and high-level control run on an NVIDIA Jetson Orin NX. The policy is executed at 50~Hz, while the low-level controller runs at 200~Hz.

\begin{table}
    \centering
    \renewcommand{\arraystretch}{1.3}
    \setlength{\tabcolsep}{7pt}
    \caption{Real-world Experiments}
    \label{tab:real_result}
    \begin{tabular}{lcccc}
        \toprule
        \textbf{Disturbance} & $E_\text{succ}\uparrow$ & $E_\text{drift}\downarrow$ & $E_\text{time}\downarrow$ & $E_\text{smth}\downarrow$ \\
        \midrule
        \rowcolor{gray!10}
        (a) \textbf{None} & & & & \\
        \hdashline
        \textbf{Ours}           & $\mathbf{10/10}$ & $\mathbf{0.532}$ & $\mathbf{1.286}$ & $\mathbf{0.266}$ \\
        Proportional Controller & 8/10  & 1.121 & 1.457 & 0.367 \\
        \hline
        \rowcolor{gray!10}
        (b) \textbf{Wind} & & & & \\
        \hdashline
        \textbf{Ours}                    & $\mathbf{8/10}$ & $\mathbf{0.839}$ & $\mathbf{1.243}$ & 0.370 \\
        Proportional Controller & 7/10  & 1.638 & 1.824 & $\mathbf{0.358}$ \\
        \hline
        \rowcolor{gray!10}
        (c) \textbf{Additional Weight} & & & & \\
        \hdashline
        \textbf{Ours}                    & $\mathbf{8/10}$ & $\mathbf{0.858}$ & $\mathbf{1.426}$ & $\mathbf{0.304}$ \\
        Proportional Controller & 4/10 & 1.797 & 1.691 & 0.364 \\
        \bottomrule
    \end{tabular}
    \vspace{-10pt}
\end{table}

\addtolength{\textheight}{-1cm}

\subsection{Experimental Results}\label{sec:real_exp_results}
We first conduct experiments in an indoor environment. To evaluate the contribution of the low-level controller to sim-to-real transfer, we use the proportional-controller variant as a baseline. To assess robustness to unknown disturbances and model mismatch, we additionally perform experiments under wind disturbance and with an additional payload. Because recovery difficulty differs between upright and inverted initial attitudes, each setting consists of 10 trials, including 5 upright starts and 5 inverted starts.

We use the same metrics as in~\Cref{sec:sim_exp_setup}, except that in the real-world experiments, $E_\text{succ}$ denotes the number of successful recoveries without a crash. For the remaining metrics, we report the mean values over successful trials only. The ground-truth quadrotor position is measured using a NOKOV motion-capture system\footnote{https://www.nokov.com}.

As shown in~\Cref{fig:indoor_exp}, the policy achieves zero-shot sim-to-real transfer and successfully recovers the quadrotor from different initial attitudes to a stable hover while maintaining limited drift. The statistical results in~\Cref{tab:real_result} show that the proposed method achieves performance comparable to that observed in simulation and consistently outperforms the proportional-controller baseline. This result suggests that the INDI controller compensates for part of the model mismatch, thereby reducing the sim-to-real gap. The proposed method also demonstrates stronger robustness under wind disturbance and with an additional payload, as shown in~\Cref{fig:wind} and~\Cref{fig:weight}. Among all metrics, the largest discrepancy between simulation and real-world deployment is observed in the drift distance. We attribute this gap primarily to the optical-flow sensor, whose measurements can become highly unreliable under certain real-world conditions

\begin{figure}
    \centering
    \includegraphics[width=1.0\linewidth]{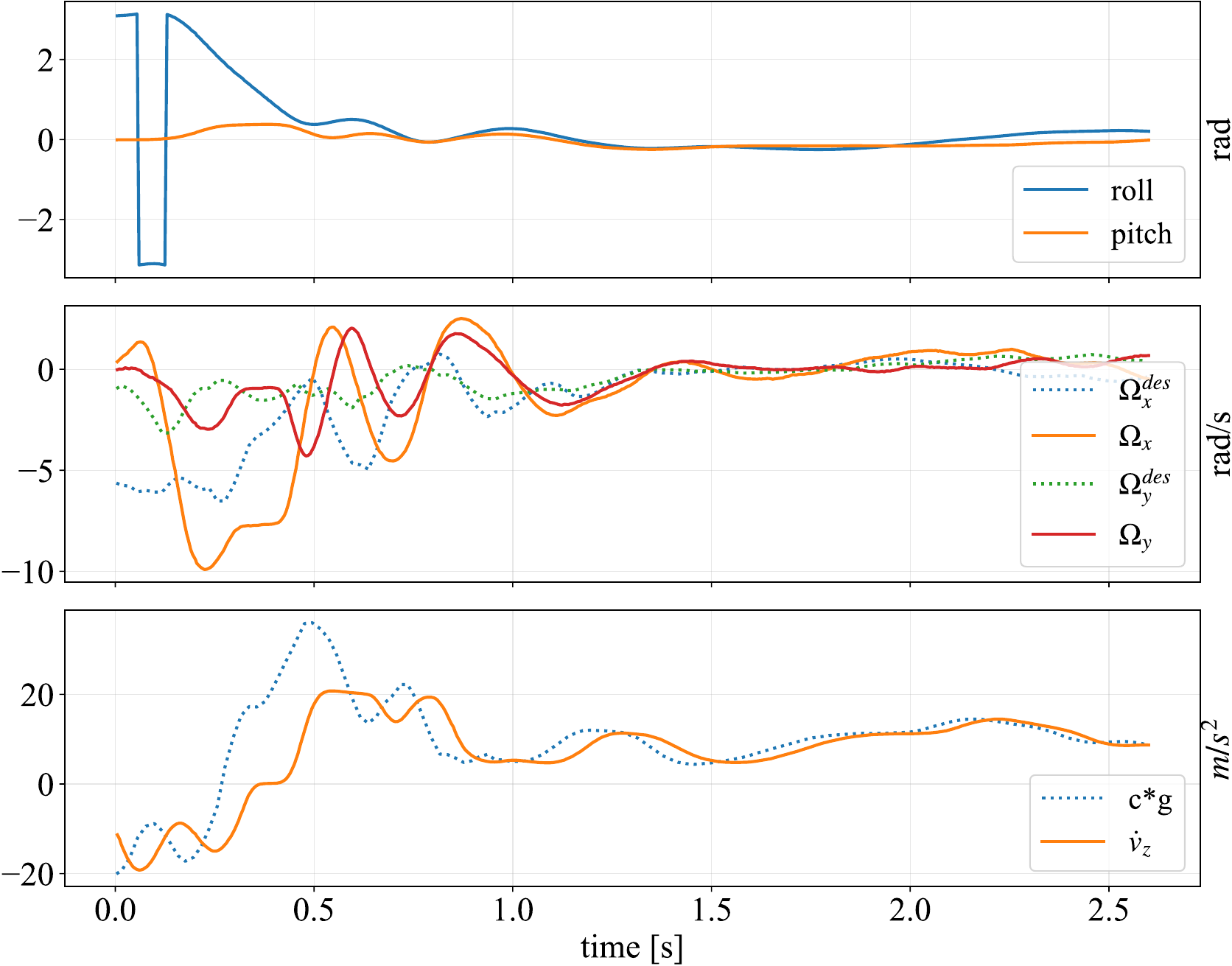}
    \caption{Statistical analysis of an outdoor experiment involving inverted takeoff. The quadrotor makes an aggressive attitude correction while utilizing reverse thrust to leave the ground. It then stabilizes its attitude and achieves a stable hover.}
    \label{fig:real_exp_data}
    \vspace{-10pt}
\end{figure}

To further examine the robustness of the learned policy, we also conduct experiments in outdoor environments. As shown in~\Cref{fig:outdoor_exp} and~\Cref{fig:real_exp_data}, the policy successfully recovers the quadrotor from different initial attitudes on a variety of ground surfaces. Notably, the policy also generalizes to takeoff from a sloped surface, even though this scenario is not explicitly included during simulation training. We suggest that the policy captures the core dynamics of fall recovery and is therefore less sensitive to variations in the surrounding environment.

\section{CONCLUSION}

In this letter, we present an RL-based framework for autonomous fall recovery of bidirectional-thrust quadrotors from arbitrary resting attitudes on the ground to a stable hover. The proposed system operates using only limited onboard sensing, without requiring explicit state estimation throughout the full maneuver. Extensive simulation experiments show that each major design choice contributes to performance. Real-world experiments further demonstrate successful zero-shot transfer and robust recovery. These results indicate that agile quadrotor fall recovery is feasible even under noisy, partial, and intermittently invalid sensing.

A limitation of the current system is that recovery performance remains sensitive to the reliability of the optical-flow measurements, which can degrade significantly in low-texture or poorly illuminated environments. In future work, we plan to investigate more robust multimodal sensing and learning strategies, as well as recovery in more cluttered environments and on more complex terrain.


\bibliographystyle{ieeetr}
\bibliography{ref}

\end{document}